%% file: nips_2018.tex
\documentclass{article}

\PassOptionsToPackage{numbers, compress}{natbib}

\usepackage[final]{nips_2018}

\usepackage[utf8]{inputenc} 
\usepackage[T1]{fontenc}    
\usepackage{hyperref}       
\usepackage{url}            
\usepackage{booktabs}       
\usepackage{amsfonts}       
\usepackage{nicefrac}       
\usepackage{microtype}      
\usepackage[skip=7pt]{caption}
\usepackage{multicol}
\usepackage{amsmath}
\usepackage{amsfonts}
\usepackage{scalerel}
\usepackage{graphicx}
\usepackage{microtype}
\usepackage{enumerate}
\usepackage{color}
\usepackage{tikz}
\usepackage{array}
\usepackage{float}
\usetikzlibrary{bayesnet}

\DeclareMathOperator*{\argmin}{argmin}

\newcommand{\E}[1]{{\mathbb E}\left[ #1 \right]}
\newcommand{\Hdg} [1] {\vspace{.5em} \noindent \textbf{#1}: }

\newcommand{\Space} [1] {\ensuremath{ \mathcal{#1}}}
\newcommand{\Aspace} [1]{\ensuremath{\Space{A}_{#1}}}
\newcommand{\Y} {\ensuremath{y(t)}}

\newcommand{\Yappk} [1]{\ensuremath{\tilde{y}_{#1}(t)}}
\newcommand{\Z} {\ensuremath{z}}

\newcommand{\As} {\ensuremath{\Aspace s}}
\newcommand{\Ap} {\ensuremath{\Aspace p}}
\newcommand{\as} {\ensuremath{a_s}}
\newcommand{\ap} {\ensuremath{a_p}}
\newcommand{\apopt} {\ensuremath{a_p^*}}
\newcommand{\param} {\ensuremath{\mathbb\theta}}
\newcommand{\PARAM} {\ensuremath{\Theta}}

\newcommand{\Oparam}[1] {\ensuremath{\mathbb\theta^{*,#1}}}
\newcommand{\Eps}  {\ensuremath{\epsilon}}
\newcommand{\Ds} {\ensuremath{\Delta}}
\newcommand{\Dsj} [1] {\ensuremath{\Delta_{#1}}}
\newcommand{\Dsk} [1] {\ensuremath{\Delta^{#1}}}
\newcommand{\Dsjk} [2] {\ensuremath{\Delta_{#2}^{#1}}}
\newcommand{\DefInt} [4] {\ensuremath{\int\limits_{#1}^{#2} #3\; \mathrm{d}{#4}}}

\newcommand{\Ln}  {\ensuremath{\hat{L}_n(\param)}}

\newcommand{\PDTNmu}  {\ensuremath{\mu(\param)}}
\newcommand{\PDTNd} {\ensuremath{\sigma_n(\param)}}

\newcommand\tuple[2]{\ensuremath{\left(#1, #2\right)}}

\title{To Stir or Not to Stir: Online Estimation of Liquid Properties for Pouring Actions}

\newcommand{\specialcell}[2][c]{%
  \begin{tabular}[#1]{@{}c@{}}#2\end{tabular}}
  
\author{
  \specialcell{Tatiana Lopez-Guevara$^{1,2}$, Rita Pucci$^{1}$, Nicholas Taylor$^{2}$,\\Michael U. Gutmann$^{1}$, Subramanian Ramamoorthy$^{1}$, Kartic Subr$^{1}$} \\
  $^{1}$University of Edinburgh, $^{2}$Heriot-Watt University\\
  \texttt{t.l.guevara@ed.ac.uk} \\
}

\begin{document}

\maketitle

\begin{abstract}
Our brains are able to exploit coarse physical models of fluids to solve everyday manipulation tasks. There has been considerable interest in developing such a capability in robots so that they can autonomously manipulate fluids adapting to different conditions. In this paper, we investigate the problem of adaptation to liquids with different characteristics. We develop a simple calibration task (stirring with a stick) that enables rapid inference of the parameters of the liquid from RBG data. We perform the inference in the space of simulation parameters rather than on physically accurate parameters. This facilitates prediction and optimization tasks since the inferred parameters may be fed directly to the simulator. We demonstrate that our ``stirring'' learner performs better than when the robot is calibrated with pouring actions. We show that our method is able to infer properties of three different liquids -- water, glycerin and gel -- and present experimental results by executing stirring and pouring actions on a UR10. We believe that decoupling of the training actions from the goal task is an important step towards simple, autonomous learning of the behavior of different fluids in unstructured environments.
\end{abstract}


\section{Introduction}
\label{sec:intro}
\input{introduction}

\begin{figure}
 \centering
 \includegraphics[width=1\linewidth]{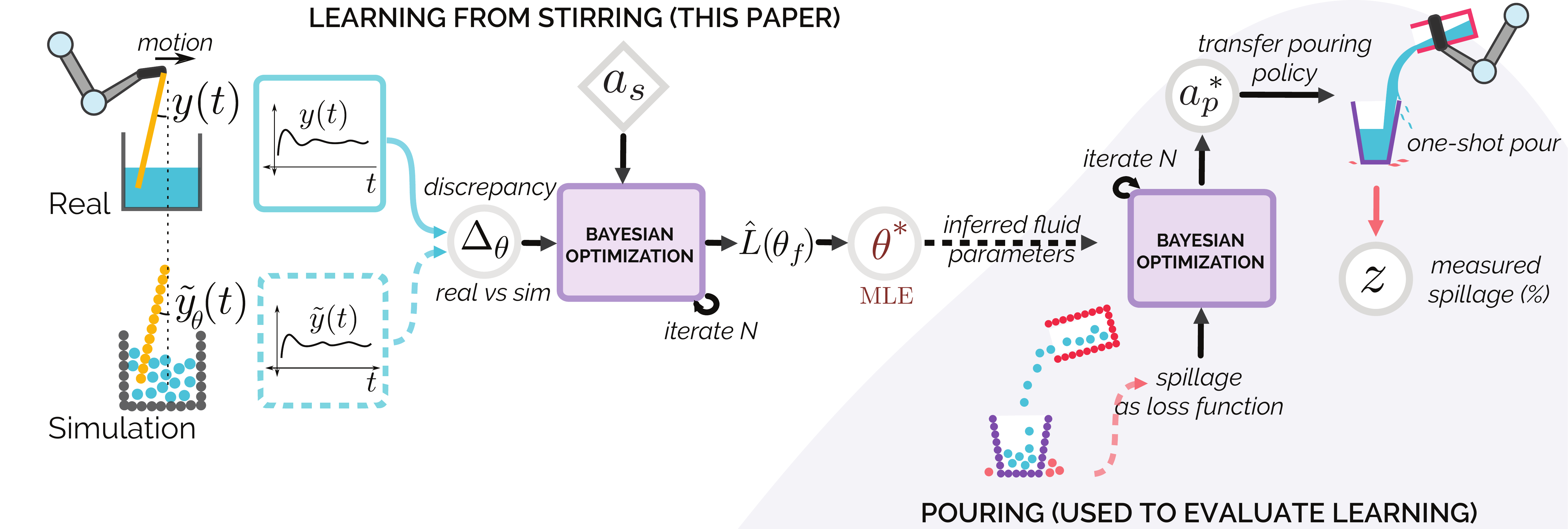}
 \caption{\label{fig:teaser} Learning parameters of liquids $\theta$ by stirring with the motion pattern $a_s$. The discrepancy $\Delta_\theta$ between the real $y(t)$ and simulated $\tilde{y}(t)$ time signals is obtained in real time.  The efficacy of learning is evaluated by executing one-shot pouring and measuring the percentage spillage \Z.}
 \vspace{-0.8em}
\end{figure}

\section{Related Work and Contribution}
\label{sec:relwork}
\input{relwork}

\section{Problem Definition}
\label{sec:problem}
\input{calib}

\section{Experiments and Results}
\input{expts}

\section{Conclusion} 
\label{sec:conclusion}

We have presented the first supervised learning algorithm for robotic manipulation of liquids that decouples the training action (stirring) from the final task (pouring) while adapting to liquids with widely different properties. Learning by stirring is preferable to learning by pouring because it is easy to automate, it is time efficient and avoids the mess involved due to spillage. We demonstrated that stirring leads to reduced spillage for water compared to state of the art and also presented results for adapting the pouring to other liquids. We discussed the several design decisions involved, along with quantitative justification and recommendations for prospective use-cases. 

\bibliographystyle{plain}
\bibliography{references}

\end{document}

%% file: introduction.tex
Empowering robots with a capability to autonomously manipulate liquids will lead to impact across sectors such as engineering and the service industry. Recent approaches have focused on learning how to pour by reasoning using simulations of the liquid~\cite{Schenck2017, CoRL17} or optimization based on parametric assumptions (parabolic trajectory) of the liquid~\cite{Pan2016_heuristic}. The physical parameters of the setup play an important role in both approaches. This includes the shapes of the pouring and receiving containers, intrinsic and extrinsic properties of the liquid, etc. Prior works have focused on inferring specific subsets of these parameters via sampling~\cite{Schenck2017} or by feedback in closed-loop~\cite{Schenck2017_2}.

A general class of methods, popularly known as \textit{intuitive physics}~\cite{battaglia2013simulation,bates2015humans,wu2016physics}, argues that coarse representations of physical processes are sufficient for many prediction tasks. Inspired by this approach, Lopez-Guevara et al~\cite{CoRL17} used an approximate (but real-time) fluid simulator NVIDIA Flex to represent the behavior of liquids. Although this enabled fast prediction, their inference problem involves mapping real world observations to the parameter space of the approximate simulator via a cumbersome calibration step involving pouring.


In this paper, we focus on the problem of inferring the behavior of different types of liquids using simple training interactions and their observed effects. Rather than learning physical properties, we parameterize liquids according to inputs specified to NVIDIA Flex. The learning algorithm searches this parameter space online. For this, the robot stirs the liquid using a \textit{motion pattern} and seeks simulation parameters that match the inclination of the stick in simulation against its observed values. Finally, we use the inferred parameters to predict the optimized pouring action for a given liquid and verify that it reduces spillage. The high-level contributions of this paper, in the context of pouring liquids, are that we: (1) decouple actions performed during training from the goal; (2) propose an online, autonomous calibration action;
 (3) achieve adaptability to different liquids.

%% file: relwork.tex
\Hdg{Estimation of physical properties}
A few approaches focus on estimating physical parameters such as volume~\cite{Do2016} and viscosity~\cite{elbrechter2015discriminating,saal2010active}, These methods exploit special measurement equipment such as RGBD cameras or tactile sensors for parameter estimation. In our context, knowledge of the physical parameters would only be useful if a high-fidelity simulation is used to optimize decision-making during manipulation of fluids. To remain practical, it is necessary to resort to approximate simulators which typically face \emph{model mismatch} since their input parameters do not coincide exactly with physical attributes such as viscosity. 
Different approaches have been proposed to learn simulator parameters from data \cite{wu2016physics,Lintusaari2017,Gutmann2016a}. Different to \citep{wu2016physics}, we do not assume a Gaussian likelihood of the real observations given the simulated ones. Rather, we learn a model of the discrepancy between real and simulated data and use it to accelerate the search using Bayesian Optimization\cite{Gutmann2016a}.

\Hdg{Robots interacting with fluids} 
Existing methods that can reason about fluids, use only simulations~\cite{Yamaguchi2016,Kunze2015,Pan2017} or a combination of simulation and real observations~\cite{Yamaguchi2016-stereo}. The latter categories of approaches suffer from the problem that approximate simulations deviate over time from reality complicating the envisioned effects of robotic manipulations~\cite{Kunze2015}. An interesting solution, proposes to use simulations in closed-loop~\cite{Schenck2017_2}, by periodically projecting the simulated particles onto the real liquid tracked in image-space using thermal imaging. There have been a few solutions on the use of supervised learning~\cite{Pan2017,Sermanet2017} and Bayesian Optimization~\cite{CoRL17} for pouring liquids. 

\Hdg{Summary} We are inspired to combine promising directions of recent work that use supervision~\cite{Sermanet2017} for learning to pour from, say video of pouring actions. Simultaneously, we retain the benefits of using approximate simulation~\cite{Schenck2017,CoRL17} since it allows a generalization to a variety of manipulations. We decouple the training task from the manipulation so that it lends itself to automation and is less messy for tasks such as pouring liquids. Finally, we perform parameter estimation in the space of inputs of the approximate simulator rather than physical units. This enables us to use these parameters directly for predictive tasks by supplying them to the simulator during test execution. 

%% file: calib.tex

Let $\as\in\As$ denote actions performed in training(``stirring'').
Let $\param\in\PARAM$ define the parameters controlling the behaviour of the liquid in the simulation-based model. For each action \as\ executed by the robot, let the observable at time $t$ be \Y, the inclination of the stick used for stirring. When the same action is executed by the simulator using the parameter $\param$, let the resulting inclination be \Yappk{\param}. We define the observed discrepancy (for stirring) over the duration $T$ of action \as\ as

\begin{multicols}{2}
\noindent
\begin{equation*}
\label{eq:Discrep}
 \Dsj{\param} = \E{\DefInt{T}{}{\left(\Y-\Yappk{\param} \right)^2}{t}}  
\end{equation*}
\centering
\scalebox{0.7}{
\begin{tikzpicture}
  
  \node[obs]                               	(y) 	{$\Ds$};
  \node[obs,    right=of y]                     (yapp) 	{$\Z$};
  \node[det, above=of y, xshift=-2em] 	(as) 	{$\as$};
  \node[latent, above=of y, xshift=2.3em]		(t) 	{$\param$};    
  \node[latent, above=of yapp, xshift=2em] 	(ap) 	{$\ap$};
  
  \edge {as,t} {y} ; %
  \edge {t,ap} {yapp} ; %
  \end{tikzpicture}
  }
\end{multicols}

Let $\ap\in\Ap$ be a pouring action and let \Z\ denote the corresponding spillage (as a percentage of the poured liquid) observed when \ap\ is executed. 

\noindent
{\centering\fbox{
  \parbox{.95\linewidth}{
  Here, we analyze the problem of inferring parameters $\theta^*\in\PARAM$ of the liquid, given a space of training (stirring) actions \As\ \emph{that are different from} the space of goal (pouring) actions \Ap.  We quantify the suitability of \As\ by measuring the percentage of liquid spilled while performing optimized one-shot pouring using $\apopt\in\Ap$ obtained from \Oparam.
    }
}}

\Hdg{Assumptions}
We assume that the shapes (geometry) of the containers are available, or can be estimated using sensors. Also, we rely on the robot's estimation of its end effector pose, to synchronise simulation with reality.

\Hdg{Inference}
Given a specific \As, say stirring using a fixed motion pattern, the goal of the inference step is to estimate the best $\param^*$ in simulation such that the discrepancy $\Dsj{\param}$ is minimal. At each iteration $k$, an action $\as$ is executed by the robot and in simulation using a hypothesized parameter $\param$.
The resulting discrepancy $\Dsjk{k}{\param}$, calculated using Eq.~\ref{eq:Discrep}, together with the parameter $\param$ are provided to a Bayesian Optimizer that learns a regression of $\theta$ over $\Dsj{\param}$ using a Gaussian process.

\begin{multicols}{2}
\noindent
    \begin{equation*}
            \begin{aligned}
            \Dsk{k}(\param) &\sim \mathcal{GP}(\mu^k(\theta), \kappa^k(\theta,     \theta^{'})) \\
            \Oparam{k} &= \argmin_{\param\in\Theta}\mu^{k}(\param)
            \end{aligned}
        \label{eq:dfdx}
    \end{equation*}
    \begin{equation*}
        \Ln \propto \Phi\left(\frac{\Eps-\PDTNmu}{\PDTNd}\right)
    \label{eq:dfdy}
    \end{equation*}
\end{multicols}

An approximation of the likelihood \cite{Gutmann2016a} can be computed using the cdf of the standard Normal distribution $\Phi$ and a threshold $\epsilon$ as (visualized in Fig.~\ref{fig:results}-\textit{Middle}):



\Hdg{Evaluation}
We quantitatively evaluate the suitability of \As\ for the problem by measuring percentage spillage using an optimized action $\apopt\in\Ap$. This is due to the lack of an existing ground truth of the parameters in the simulator given its approximate nature.  Since our contribution concerns the training task, we use a pouring strategy exactly as proscribed by previous work~\cite{CoRL17}. They use a simulator to identify \apopt, by defining the loss function to be the ratio of the spilled particles to the total number of particles simulated. The $j^{th}$ iteration of their method therefore involves executing the simulator with action $\ap^j\in\As$ and $\theta^*$. The minimization results in \apopt\ after a finite number of iterations ($15$ in our case). Finally, we execute \apopt\ using the robot and measure the percentage of liquid spilled.

%% file: expts.tex
\newcommand{\Tw} {\ensuremath{\Theta_w}}
\newcommand{\paramjk} [2]{\ensuremath{\param_{#2}^{#1}}}

\subsection*{Experimental setup}
\vspace{-0.5em}
\Hdg{Stirring}
For all our experiments, we used a UR10 robot equipped with a gripper holding a stick so that it is free to pivot at the gripping point.
 Before stirring begins, the stick is vertical and partly submerged in the liquid. The motion of the end effector is limited to a plane \Space{P} parallel to the ground plane. Due to this motion, and the the resistance encountered by the stick due to the liquid, at any instant $t$, the stick might deviate from its vertical position to \Y. The inclination is intricately dependant on the velocity of the end effector and the physical properties of the liquid and the stick. \Y\ is estimated using simple computer vision on the video feed from 2 webcams with image planes orthogonal to \Space{P}. The position of the end effector of the robot is queried at $30$Hz and supplied to the simulator which replicates the executed action. The inclination produced in simulation at instant $t$ is recorded as \Yappk{\param}. 
 The space of stirring actions \As\ is discrete and determined by the stirring pattern. In this work we used a cyclic sequence, $\As=\{\as^i\}, \; i=1,\cdots, m$ that visually follows an $m-$point star with $m=9$. 

\Hdg{Pouring}
We replicate the one-shot pouring solution in existing work~\cite{CoRL17}. For completeness, we review their method here using our notation. The space of pouring actions \Ap\ is two dimensional and continuous. The 2D space is parameterized  by a constant angular velocity and the relative distance between source and target containers $\ap^i = \tuple{\omega^i}{p^i}$. After $15$ iterations of the optimizer over $\ap$ given $\param^*$, the robot obtains an estimate for the optimal pouring action \apopt, which it then executes. We measure the percentage of liquid spilled by the robot over 5 repetitions of the above experiment. 

\subsection*{Discussion}
\label{sec:discussion}
\begin{figure}
\centering
 \includegraphics[width=1\linewidth]{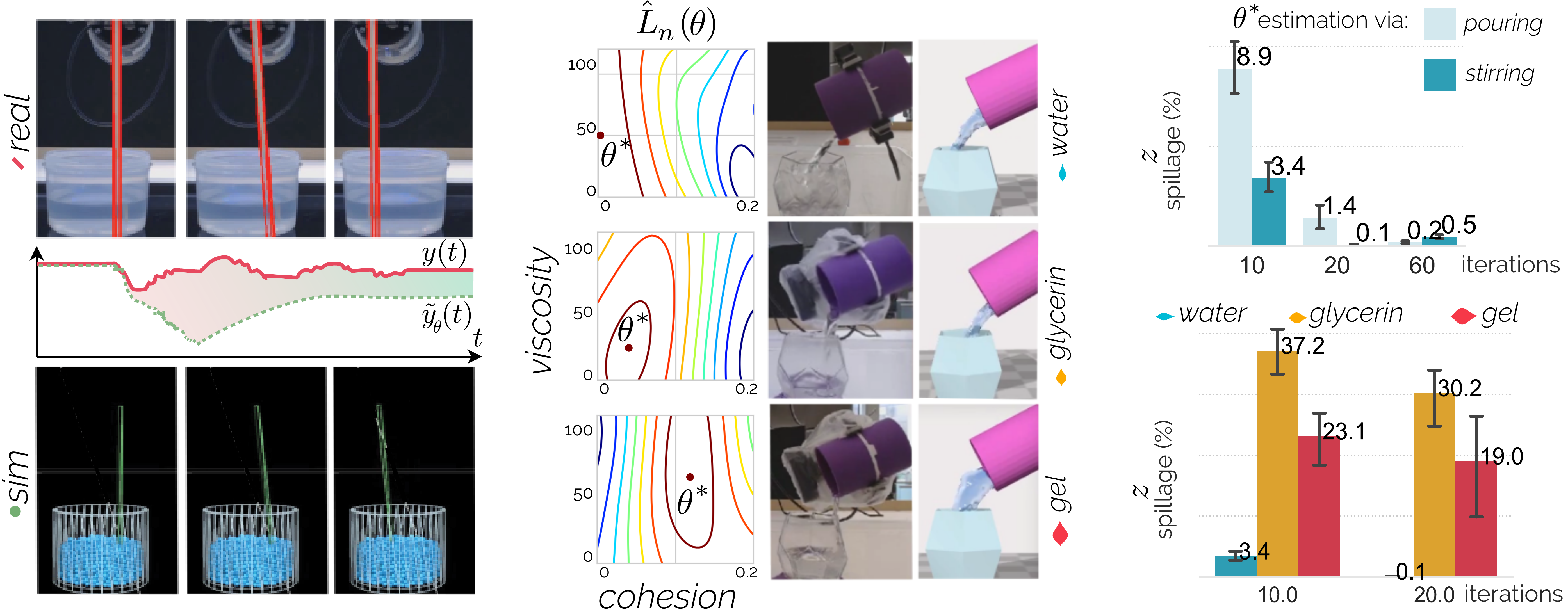}
\caption{\label{fig:results}\emph{Left:} Inclination vs time for the stick position (red) and closest sim-hypothesis (green) for one of the repetitions in water. Discrepancy is proportional to the shaded area. \emph{Middle:} Contour plots of the posterior belief on the fluid parameters after stirring water, glycerin and gel for one of the repetitions. \emph{Right-up:} Effect of two calibration methods in the deployment task measured as the decrease of spillage with respect to the number of iterations. \emph{Right-down:} Effect of the parameters inferred after performing the stirring action 10 and 20 times on three liquids.}
\vspace{-0.8em}
\end{figure}

\vspace{-0.5em}
\Hdg{Learning by stirring vs learning by pouring}
We compared the percentage spillage \Z\ achieved by our algorithm which learns by stirring against the method proposed in~\cite{CoRL17} which calibrates by pouring using a training cup. Although it would seem intuitive that applying the same task to train must result in lower spillage under test conditions, our results indicate the contrary. Fig.~\ref{fig:results}-\textit{Right-up} plots \Z\ vs $N$, where $N$ is the number of iterations of the B.O. used to estimate $\theta^*$. Using our stirring approach, the spillage is less than $5\%$ even with only 10 iterations, under half the corresponding figure when the robot was trained with pouring. At $N=20$ iterations, our approach almost achieves zeros spillage (which is lower than learning from pouring at $N=60$ iterations). 

\Hdg{Pouring other liquids}
We observed a similar trend across three different liquids (Fig.~\ref{fig:results}-\textit{Right-down}): as $N$ is increased, the spillage reduces. However, the degree of spillage is significantly higher for more glycerin and gel. On further investigation of the video and the simulator, we realized that the excessive spillage for glycerin is due to the unusually high adhesive effect that makes glycerin stick to the pouring container. Unfortunately, this adhesive behaviour cannot be modelled by the simulator on a particle-particle interaction.  We conclude that the choice of the approximate simulator, combined with potentially different behavior across training and pouring actions might be a source of error during spillage. However, the capability to infer parameters within a limited gamut of expressibility is still a valuable addition to the toolkits proposed by existing methods.
